\documentclass[letterpaper, 10 pt, conference]{ieeeconf}  %
\IEEEoverridecommandlockouts                              %

\usepackage{flushend}
\usepackage{lipsum}
\usepackage[utf8]{inputenc}
\usepackage{graphicx}
\usepackage{color}
\usepackage{latexsym}
\usepackage{bmpsize}
\usepackage{xurl}
\usepackage{comment}
\usepackage{amsmath,amssymb}
\usepackage{mathtools}  %
\usepackage{siunitx}
\usepackage[nobreak]{cite}
\usepackage[subrefformat=parens]{subcaption}
\usepackage[hyperfootnotes=false]{hyperref}
\usepackage{cleveref} %
\usepackage{here}
\usepackage{listings} %
\usepackage[affil-it]{authblk} %
\usepackage{todonotes}

\crefname{figure}{Fig.}{Fig.}
\crefname{table}{Table}{Table}
\crefname{section}{Section}{Section}
\renewcommand{\todo}[1]{}
\newcommand{\circletext}[1]{\raise0.2ex\hbox{\textcircled{\scriptsize{#1}}}}

\def\Underline{\setbox0\hbox\bgroup\let\\\endUnderline}
\def\endUnderline{\vphantom{y}\egroup\smash{\underline{\box0}}\\}

\definecolor{myComment}{rgb}{0.0, 0.6, 0.0}
\definecolor{myKeyWord}{cmyk}{1.0, 0.0, 0.0, 0.3}
\definecolor{myString}{cmyk}{0.0, 1.0, 0.0, 0.0}

\lstdefinestyle{customText}{
    backgroundcolor  = {\color{white}},
    basicstyle       = {\footnotesize},
    breaklines       = {true},
    commentstyle     = {\itshape  \color{myComment}},
    keywordstyle     = {\bfseries \color{myKeyWord}},
    lineskip         = {-0.5ex},
    showstringspaces = {false},
    sensitive        = {true},
    stepnumber       = {1},
    stringstyle      = {\ttfamily \color{myString}},
    tabsize          = {2},
    xleftmargin      = {2zw},
    xrightmargin     = {2zw}
}
\begin{document}

\title{\bf{\LARGE{{Real-time Batched Distance Computation for Time-Optimal Safe Path Tracking}}}}

\author[1,2]{Shohei Fujii\thanks{E-mail: \href{mailto:SHOHEI001@e.ntu.edu.sg}{SHOHEI001@e.ntu.edu.sg}; Corresponding author} }
\author[1,3]{Quang-Cuong Pham}
\affil[1]{School of Mechanical and Aerospace Engineering, Nanyang Technological University, Singapore}
\affil[2]{DENSO CORP., Japan}
\affil[3]{Eureka Robotics, Singapore}
\maketitle

\begin{abstract}

In human-robot collaboration, there has been a trade-off relationship between the speed of collaborative robots and the safety of human workers.
In our previous paper, we introduced a time-optimal path tracking algorithm designed to maximize speed while ensuring safety for human workers~\cite{fujii2023timeoptimal}.
This algorithm runs in real-time and provides the safe and fastest control input for every cycle with respect to ISO standards~\cite{ISOTS15066}.
However, true optimality has not been achieved due to inaccurate distance computation resulting from conservative model simplification.
To attain true optimality, we require a method that can compute distances 1. at many robot configurations to examine along a trajectory 2. in real-time for online robot control 3. as precisely as possible for optimal control.
In this paper, we propose a batched, fast and precise distance checking method based on precomputed link-local SDFs.
Our method can check distances for 500 waypoints along a trajectory within less than 1 millisecond using a GPU at runtime, making it suited for time-critical robotic control.
Additionally, a neural approximation has been proposed to accelerate preprocessing by a factor of 2.
Finally, we experimentally demonstrate that our method can navigate a 6-DoF robot earlier than a geometric-primitives-based distance checker in a dynamic and collaborative environment.
\end{abstract}

\section{Introduction} \label{sec:intro}
\begin{figure}[tbp]
  \begin{center}
  \includegraphics[width=1.0\hsize]{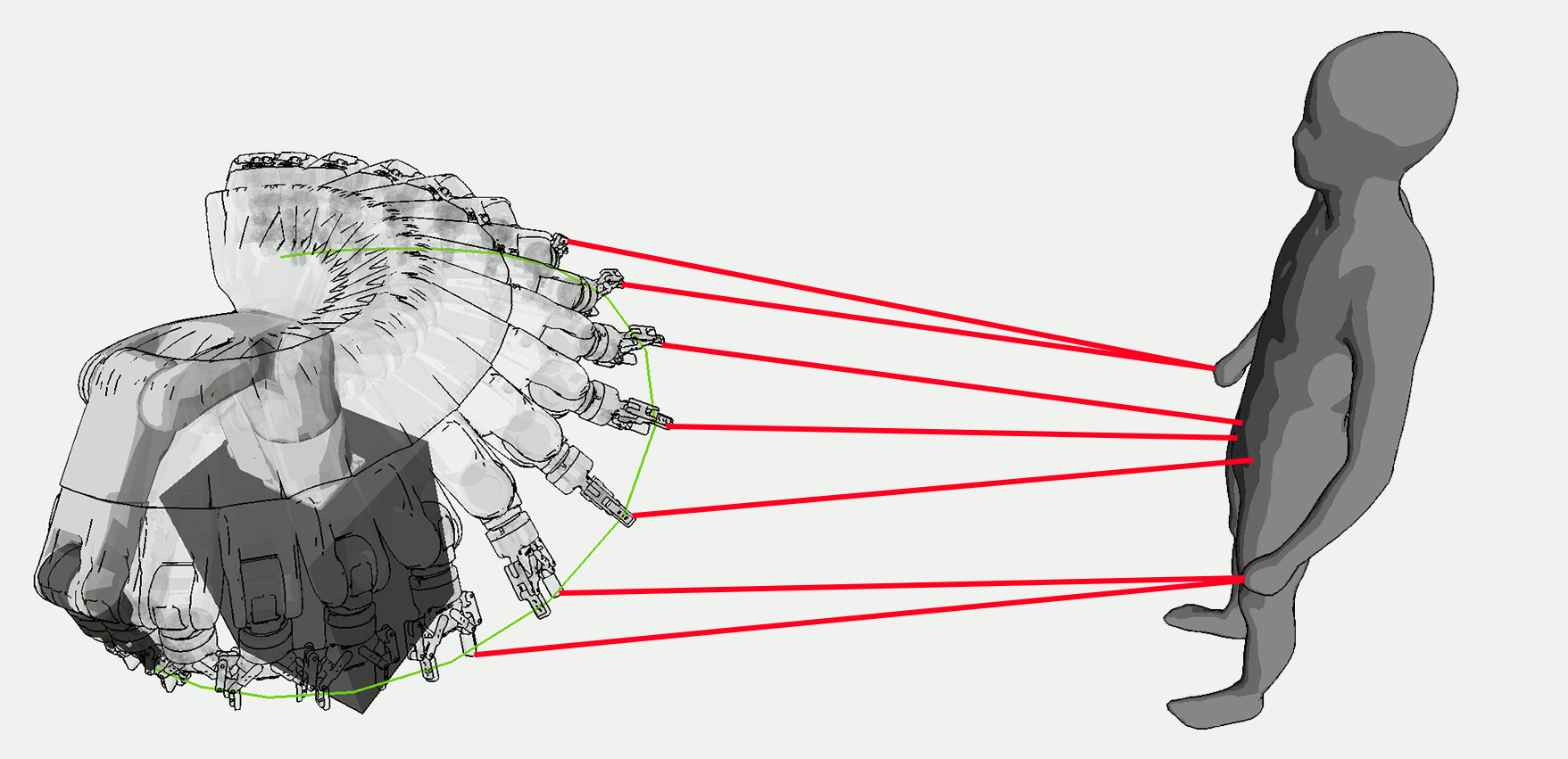}
  \caption{
   Problem Setting Overview: Computing distances in real-time, across multiple robot configurations, with precision. See \cref{sec:intro} for more information.
    }
  \vspace{-0.5cm}
  \label{fig:problem_formulation}
  \end{center}
\end{figure}

Collaborating with robots while ensuring human safety has been a critical challenge, as slowing down the robot operation to mitigate injuries will impede productivity.
To maximize the productivity of collaborative robots while guaranteeing the safety, we have proposed time-optimal path tracking algorithm~\cite{fujii2023timeoptimal} which runs in real-time and provides the safe and fastest control input with respect to \emph{Speed and Separation Monitoring} in ISO standards~\cite{ISOTS15066}.
In this path-tracking method, distances between the obstacles and a robot for waypoints along an executing trajectory must be given.
Given the distances, the algorithm computes the fastest velocity profile and navigates the robot in a time-optimal manner~(\cref{fig:problem_formulation}).
Finally, the control input is sent to a robot to follow the derived velocity profile.
This whole process must run in every control cycle, which is about 10 \si{ms} according to the communication protocol of industrial robots\footnote{For example, the control period is 8 \si{ms} in the case of DENSO b-CAP communication protocol \url{https://www.denso-wave.com/en/robot/product/function/b-CAP.html}.}.

To achieve true optimality in path tracking, the precise distances need to be given.
In our previous paper, the robot is simplified with spheres and the distances between the spheres and voxels are computed with \emph{hypot} function using their center positions.
Such distance checking with a simplified model can run almost in real-time.
However, the computed distances are smaller than its actual value due to its simplification, which makes the robot's behavior conservative and exacerbates the productivity of the robot.
In contrast, a exact mesh-to-mesh distance checker cannot run in real-time (experimentally, 130 \si{\micro \second} per one configuration, 65 \si{ms} per one trajectory with FCL~\cite{FCL}).
To the best of our knowledge, no existing distance checker is applicable to real-time safety control.

In this paper, we propose a batched, fast and precise distance checker based on pre-computed link-local Signed Distance Fields(SDFs) to address this issue.
Leveraging GPU parallelization for pre-processing of robot's SDFs, the proposed method is able to check distances for multiple robot configurations within less than 1 \si{ms} at runtime.
Additionally, a neural approximation of the pre-processing has been proposed, resulting in 2x faster pre-processing.
Finally, we experimentally demonstrate that our distance checker actually navigates a robot faster than the method using a robot modeled with spheres in a dynamic, collaborative environment.

The paper is organized as follows.
We survey related work in \cref{sec:related_work}.
\cref{sec:our_method} presents our parallel distance checking method and some techniques to reduce the pre-processing time in a constant order including the neural approximation.
In \cref{sec:experiment}, we evaluate the performance of the neural approximation and also examine that the approximation does not affect the precision of distance computation.
Then, the experimental comparison is shown for the real-time safe path tracking in a collaborative environment.
Finally, we discuss the limitations of our approach and conclude with some directions for future work in \cref{sec:conclusion}.

\section{Related Work} \label{sec:related_work}

\subsection{Model-based Distance Computation}
In collision detection and distance computation between 3D models, hierarchal data structure, or `broad-phase structure', is commonly applied to filter out object pairs that are far away and dramatically accelerates the collision/distance queries.
Examples of such data structures include AABB Tree, OCTree and Inner Sphere Tree~\cite{panRealtimeCollisionDetection2013,FCL,kaluschkeMassivelyParallelProximityQueries2014}.
However, these data structures are optimized for CPU and lack batch-processing capabilities, hence they do not have an sufficient throughput for real-time safety control.
For instance, the distance query with an octree for 1000 configurations will take $39.1$ \si{ms} according to \cite{panRealtimeCollisionDetection2013} (note that this does not include the octree construction time), which is still slow for real-time safety control. %
Another challenge is that the throughput depends on the positions of robots and obstacles, which makes it difficult to ensure its constant execution time at the time of deployment of the system.
In contrast, our approach does not depend on runtime-varying settings except for the number of configurations.

Simplification of robot/human models with geometric primitives such as spheres and capsules is commonly used for distance computation in motion planning and safe robotic operation~\cite{ratliffCHOMPGradientOptimization2009,liuAlgorithmicSafetyMeasures2016,safeeaEfficientCalculationMinimum2019,secilMinimumDistanceCalculation2022}.
However, evaluating distances for multiple configurations in a batched manner using primitives other than spheres are actually slow.
In fact, our preliminary experiment shows that distance computation between a 6 DoF robot simplified with 7 capsules and (only) 3000 points for 500 configurations took about 70 \si{ms} even on GPU, which is not applicable to real-time control.
This is primarily because the projection of points onto the axis of capsules requires a time-consuming (batched) matrix multiplication at runtime.

\subsection{Signed Distance Fields (SDFs)}
In unknown environments, prior knowledge about obstacles such as their shape, size and position is not always accessible.
Therefore, Signed Distance Fields(SDFs) or Un-signed Distance Fields(USDFs) of an \emph{environment} are often used because these does not necessarily require a prior knowledge on obstacles and offers the gradient of (U)SDFs to push the trajectory away from obstacles~\cite{ratliffCHOMPGradientOptimization2009,dongMotionPlanningProbabilistic2016,mukadamContinuoustimeGaussianProcess2018}.
There are a number of methods for SDFs construction; some come from a context of SLAM ~\cite{kinectfusion2011,oleynikovaVoxbloxIncremental3D2017,hanFIESTAFastIncremental2019} and some from that of machine learning/NeRF~\cite{ortizISDFRealTimeNeural2022}.
However, most of them assume a \emph{static} environment and incrementally construct a scene since SDFs reconstruction requires data propagation which is inherently time-consuming.
To the best of our knowledge, no previous work satisfies all the requirements for the safe-control application: `batched', `real-time' and `precise'.

One of the promising work is ~\cite{fineanSimultaneousSceneReconstruction2021} which builds on ~\cite{juelgFastOnlineCollision2017,hermannUnifiedGPUVoxel2014}.
This method computes the SDFs of an environment from an incoming sensory pointcloud in parallel on GPU using a Parallel Banding Algorithm ~\cite{caoParallelBandingAlgorithm2010}.
The distance data for the voxels occupied by the robot is then retrieved.
In their demonstration, they showcase online motion planning of a mobile manipulator platform.
However, the computation time of this method depends on the size of environment due to the data propagation and is not suitable especially for large environments.
Most importantly, their reported SDFs construction time is $17.5 \pm 0.4$ \si{ms} for 5cm resolution and $36.2 \pm 8.3$ \si{ms} for 2.5 cm resolution, which is not fast enough for real-time control %
(Note that the `SDFs computation time' does not include the time for distance queries).

Another interesting work is ReDSDF~\cite{liuRegularizedDeepSigned2022}, a machine learning based SDFs estimator that employs a neural network which takes query points and poses as inputs and outputs distances for each of them.
While its estimation accuracy is sufficient for safety-critical use-cases, their architecture is not suitable for real-time safety control due to the following reasons:
For robot's SDFs generation: 1. ReDSDF requires retraining of a neural network for any change in a robot, and 2. the neural network needs to be evaluated for each waypoint along a trajectory, repeatedly feeding the same query points. These requirements make ReDSDF unsuitable.
For environment's SDFs construction: 1. ReDSDF requires a model that is trained for each individual obstacle, and 2. each obstacle needs to be tracked in some way; these are not realistic for industrial applications.

\begin{figure}[tbhp]
  \begin{center}
  \includegraphics[width=0.7\hsize,trim={2.5cm 55cm 22cm 6cm},clip]{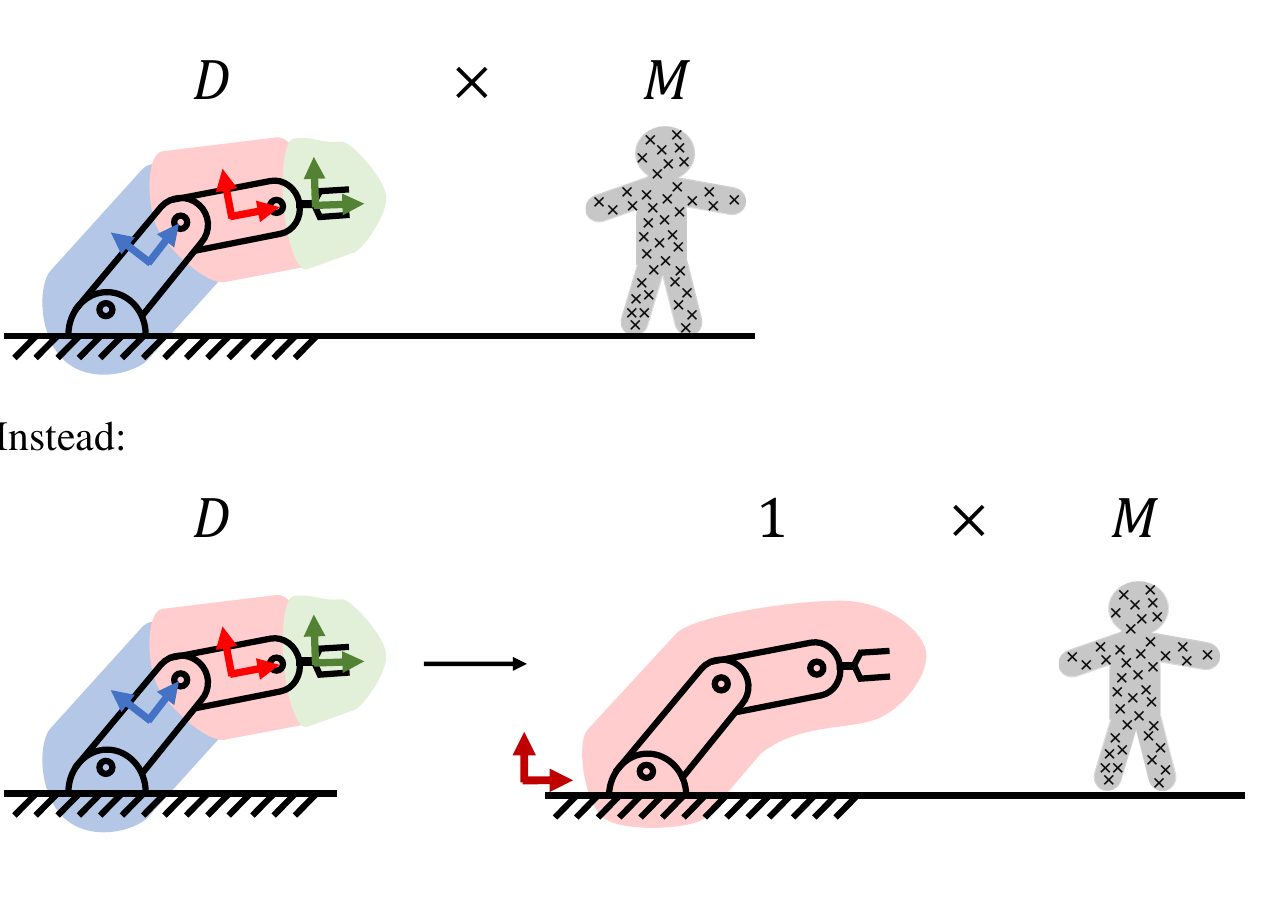}
  \caption{A common way to compute a distance between pointcloud and SDFs requires pointcloud transformations for each link coordinate. Instead, we transform and merge the link-local SDFs into the global coordinates in a pre-processing stage, and then evaluate it to obtain distances at runtime.}
  \label{fig:design_policy}
  \end{center}
\end{figure}

Some previous methods employ link-local SDFs of a \emph{robot}, instead of an \emph{environment}, for distance computation in motion planning or collision avoidance~\cite{hauserSemiinfiniteProgrammingTrajectory2021,xuObstacleAvoidanceManipulator2018}.
In these approaches, the pointcloud is transformed into every link coordinate and the distance is then obtained from link-local SDFs, resulting in a computational complexity of $O(DM)$ where $D$ is the DoF of a robot and $M$ is the number of pointcloud.
This computation heavily depends on the number of points and its batch processing is often insufficiently fast for real-time robot control.
In contrast, our method performs the transformation of the link-local SDFs onto the coordinates of the environment beforehand, eliminating the need for pointcloud transformations (\cref{fig:design_policy}).
This leads to a faster distance retrieval in the real-time distance computation phase.

\section{Batched Robot SDFs Computation}\label{sec:our_method}

\subsection{Overview}
\begin{figure*}[thbp]
  \begin{center}
  \includegraphics[width=0.8\textwidth,trim={1.8cm 21cm 0cm 0cm},clip]{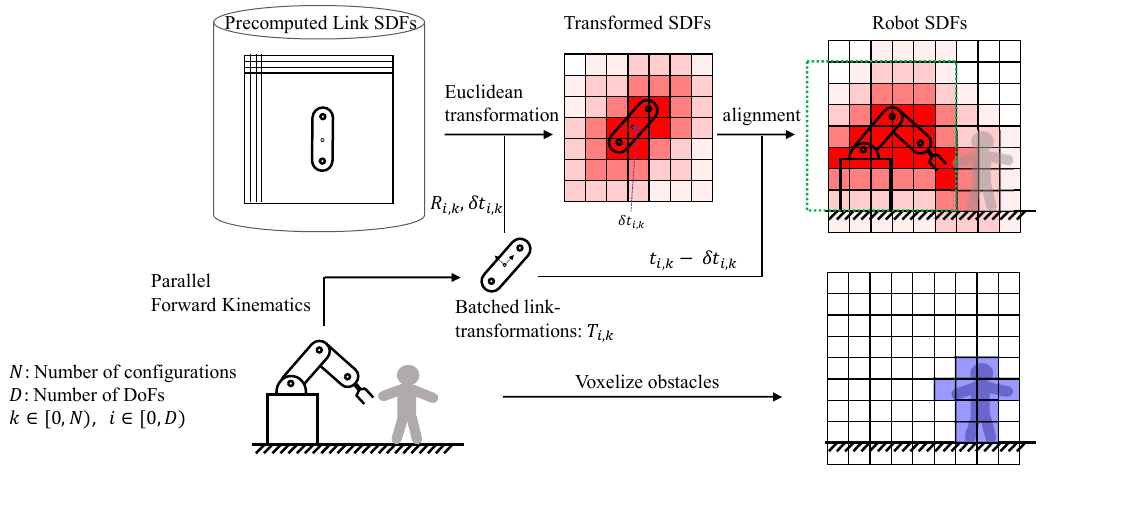}
\caption{A pipeline of parallel batched distance checking with pre-computed link-wise signed distance fields (SDFs). This is in 2D for clear illustration, but actual computation is in 3D and in a batched manner.
  See \cref{sec:our_method} for detail.}
\label{fig:method_detail}
\end{center}
\vspace*{-3mm}
\end{figure*}

The pipeline of our parallel distance checking is illustrated in \cref{fig:method_detail}.
We consider a $D$-DoF robot and examine distances at $C$ robot configurations ($\mathbf{\theta_{c}} \in \Theta$).
The environment is discretized into voxels whose extent is $\mathbf{e_e} = (e_{ex}, e_{ey}, e_{ez})$ and resolution is $\mathbf{r_e} = (r_{ex}, r_{ey}, r_{ez})$.
The total number of voxels of the environment $V_e$ is $\prod{\frac{2 \mathbf{e_e}}{\mathbf{r_e}}}$.

At the preprocessing stage, given a robot model, we pre-compute Signed Distance Fields(SDFs) for each link on its local coordinates.
We call it as \emph{Link SDFs}.
We refer $\mathbf{e_r}$ to the extent of Link SDFs $(e_{rx}, e_{ry}, e_{rz})$ and $\mathbf{r_r}$ to the resolution of Link SDFs $(r_{rx}, r_{ry}, r_{rz})$.
The size of the precomputed Link SDFs, $2\mathbf{e_r}$, must be divided by the resolution of the voxelized environment $\mathbf{r_e}$ without residue for alignment operation that is later introduced.
The resolution of Link SDFs is arbitrary and recommended to be finely voxelized.

Next, given the configurations $c$, we compute transformations $T_{i,c}$ of each link $i$ by applying parallel forward kinematics. %
Then, according to $T_{i,c}$, Link SDFs are transformed and aligned into the voxels of the environment.
We call the first euclidean transformation operation as ``euclidean transformation'' and the second alignment operation as ``alignment''.
To compute Robot SDFs for each configuration, a minimum value of the transformed Link SDFs for each link and for each voxel is taken.
Besides, obstacles in the environment are voxelized.
By extracting the distances at the voxels occupied by the obstacles and taking the minimum value for each link, the distance between the robot and obstacles can be computed.

More specifically, at the ``euclidean transformation'' stage, we shift the rotated Link SDFs within the half range of the voxel by $\delta t_{i,c} \in (-\frac{r_e}{2}, \frac{r_e}{2})$.
And then, at the ``alignment'' stage, we translate the transformed SDFs by $t_{i,k} - \delta t_{i,k}$ and snap them into the environment voxels.
The shift operation is necessary for exact alignment since the position of each link in the environment is not usually at the exact center of the voxel.
The transformation can be computed in the scheme of affine grid transformations~\cite{jaderbergSpatialTransformerNetworks2016} ~\footnote{Please refer to pytorch's documentation as well: ~\url{https://pytorch.org/docs/stable/generated/torch.nn.functional.affine_grid.html}}.
$\delta t_{i,c}$ can be computed by following the simple equations:
\begin{align}
  T_{Oi,c}       & = T_{i,c} - (- e_e)                                          \\
  k_{i,c}        & = \lfloor T_{Oi,c} / r_e \rfloor - \lfloor e_e / e_r \rfloor \\
  \delta t_{i,c} & = T_{Oi,c} - (k_{i,c} \cdot r_e + e_r)
\end{align}
where $\mathbf{e_e}$ is the 3D extent of environment, $\mathbf{r_e}$ is the 3D resolution of environment, and $\mathbf{e_r}$ is the 3D extent of Link SDFs.
The total number of voxels in transformed SDFs $V_r$ is $\prod{\frac{\mathbf{e_r}}{\mathbf{r_e}}}$.

At runtime, to compute distances against obstacles in the environment based on the Robot SDFs, we voxelize the obstacles and extract the values from Robot SDFs that is occupied by the voxelized obstacles.
By reducing the extracted values with \emph{min} for each configuration $c$, we can obtain minimum distance between a robot and the obstacles for each $c$.
This process is fast because it only reads the data on the GPU memory and does not require any calculation.

As an extra bonus, self-collision detection can be done by aligning ``in a predefined and alternating the order of checking, paying attention to the robot\'s kinematics'' as in \cite{hermannGPUbasedRealtimeCollision2013}, though it is not applied in our experiment since self-collision is usually examined in the motion-planning phase rather than the execution phase.

\subsection{Techniques for computation time reduction in a constant order}
We introduce the following techniques to optimize the computation time in a constant order.

\subsubsection{Euclidean Grid Approximation with A Tiny Neural Network}
The computation of euclidean grid transformation mapping is mathematically a matrix multiplication.
Given the center positions of grids $p_{j, xyz} \text{for } j \in [0, V_r)$ where $V_r$ is the number of grids in each Link SDFs
and link transformations $T_{i,c}$, euclidean grid transformations $G_{i,c}$ can be computed as follows:
\begin{equation}
  \begin{aligned}
    P_{xyz} \vcentcolon=
    \begin{pmatrix}
      \cdots & p_{j, xyz} & \cdots
    \end{pmatrix}                                               \\
    \begin{pmatrix}
      G_{i,c} \\
      1
    \end{pmatrix} = \begin{pmatrix}
                      R_{i,c} & \frac{\delta t_{i,c}}{e_r} \\
                      0       & 1
                    \end{pmatrix}^{-1} \begin{pmatrix}
                                         P_{xyz} \\
                                         1
                                       \end{pmatrix} \\
    = \begin{pmatrix}
        R_{i,c}^T & - R_{i,c}^T \frac{\delta t_{i,c}}{e_r} \\
        0         & 1
      \end{pmatrix} \begin{pmatrix}
                      P_{xyz} \\
                      1
                    \end{pmatrix}
  \end{aligned}
\end{equation}
Note that $p_{i, xyz}$ are normalized in the range of $[-1, 1]$ with the voxel resolution of environment voxels $\mathbf{e_r}$.
The number of column of $P_{xyz}$ is the total number of voxels in the transformed Link SDFs.
This batch processing of matrix-matrix multiplication is actually time-consuming, because general matrix-matrix multiplications of BLAS libraries provided by vendors are optimized for square matrices and we cannot leverage full performance of dedicated devices including GPU for tall-and-skinny matrices~\cite{ernstPerformanceEngineeringReal2021}.
Instead, we use a tiny neural network $f$ which is composed of two fully-connected layers and one ReLU activation layer to approximate and simplify this operation:
\begin{equation}
  \begin{aligned}
    \delta t_{inv\ i,c} = - R_{i,c}^T \frac{\delta t_{i,c}}{e_r} \\
    G_{i,c} = f(R_{i,c}) + \delta t_{inv\ i,c}
  \end{aligned}
\end{equation}
$f$ takes a rotation matrix and output euclidean grid transformations only for the specified rotation.
Recent advance in deep learning provides us a highly-optimized API for neural approximation \footnote{See \url{https://developer.nvidia.com/cudnn}}.
Since the original operation is deterministic and robot-model agonistic, we can train the neural network quickly (about 10--20 mins) and reuse the pre-trained models for any type of robots once it is trained without any additional training.
As described in \cref{subsec:neuralnet_approximation}, the maximum error of this approximation is about 1mm in our setting, which is covered by the discretization error of SDFs and therefore negligible.
Therefore, the maximum error from the ground truth which derives from the total pipeline is only the discretization error: $\frac{|r_e|}{2} + \frac{|r_r|}{2}$.

\subsubsection{Grids in Sphere instead of using Cubic Grids}
Another small technique to reduce computation time is to compute transformed SDFs only for the grids in a sphere of a radius $e_r + sqrt(3(\frac{r_e}{2})^2)$ which inscribes the link.
We can roughly reduce the number of grids by: $ \frac{\frac{4}{3}\pi e_r^3 }{(2e_r)^3} \approx 0.53 $.

\section{Experiments and Results} \label{sec:experiment}

\subsection{System setup} \label{subsec:system_setup}
All the experiments are done on a single machine, on which AMD Ryzen\textsuperscript{\texttrademark} 9 4900HS and
NVIDIA GeForce RTX\textsuperscript{\texttrademark} 2060 with Max-Q Design are equipped for CPU and GPU.
We use PyTorch and develop custom CUDA kernels for evaluation. %

\subsection{Precision and Speed of Neural Approximation for Euclidean Grid Transformations} \label{subsec:neuralnet_approximation}
\begin{figure}[thbp]
  \begin{center}
  \begin{minipage}[b]{0.99\linewidth}
    \centering
    \includegraphics[width=0.9\hsize,trim={0cm 1cm 0cm 1cm},clip]{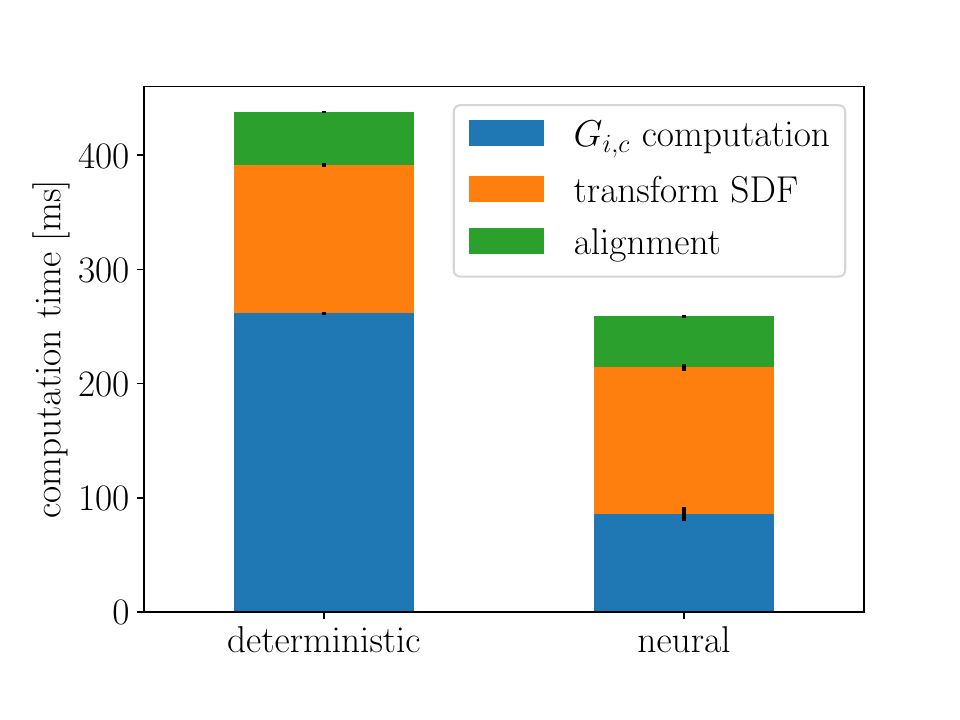}
    \subcaption{}
    \label{fig:neuralapproximationspeed}
  \end{minipage}
  \caption{Computation speed of euclidean grid approximation by a tiny neural net. %
  See \cref{subsec:neuralnet_approximation} for detail.}
  \label{fig:neuralnet_approximation_experiment}
  \end{center}
  \vspace{-3mm}
\end{figure}

First, we examine the effect of approximation of euclidean grid transformations in \cref{fig:neuralnet_approximation_experiment}.
In this experiment, we train the tiny neural network of 32 hidden parameters %
using L1 loss and Adam optimizer with a learning rate of $1e^{-4}$.
We measured the time to compute euclidean grid transformations for 500 configurations of 6 DoF robot.
Deterministic transformations are operated by a function `torch.matmul' which internally uses cuBLAS's sgemm~\footnote{\url{https://developer.nvidia.com/cublas}}.

We compare the computation speed in \cref{fig:neuralapproximationspeed}.
As a result, the neural approximation of euclidean transformations $G$ is about 3.2x faster than the deterministic one, and the total SDFs computation becomes about 2x faster.
According to Figure 22 of \cite{ernstPerformanceEngineeringReal2021}, ours (3.2x) exceeds the performance gain of \cite{ernstPerformanceEngineeringReal2021} (almost 2x) from cuBLAS.

We also test approximation errors from the ground truth.%
We use $10^7$ randomly-generated link transformations to examine the maximum error.
The result is that the approximation error of the euclidean grid transformations is less than 0.0013 at the maximum.
This means that, in the following experiment, considering the extent of Link SDFs $e_r$ is set to $1.2$ \si{m}, the actual error in $G$ is $0.0013 \times e_r = 1.56$ \si{mm} which is way smaller than the grid discretization size (1 \si{cm}) and therefore negligible.

\subsection{Comparison with a robot in simulation} \label{subsec:sim_robot_comparison}

\begin{figure}[thbp]
\begin{center}
  \includegraphics[width=0.80\hsize,trim={2cm 2cm 2cm 1cm},clip]{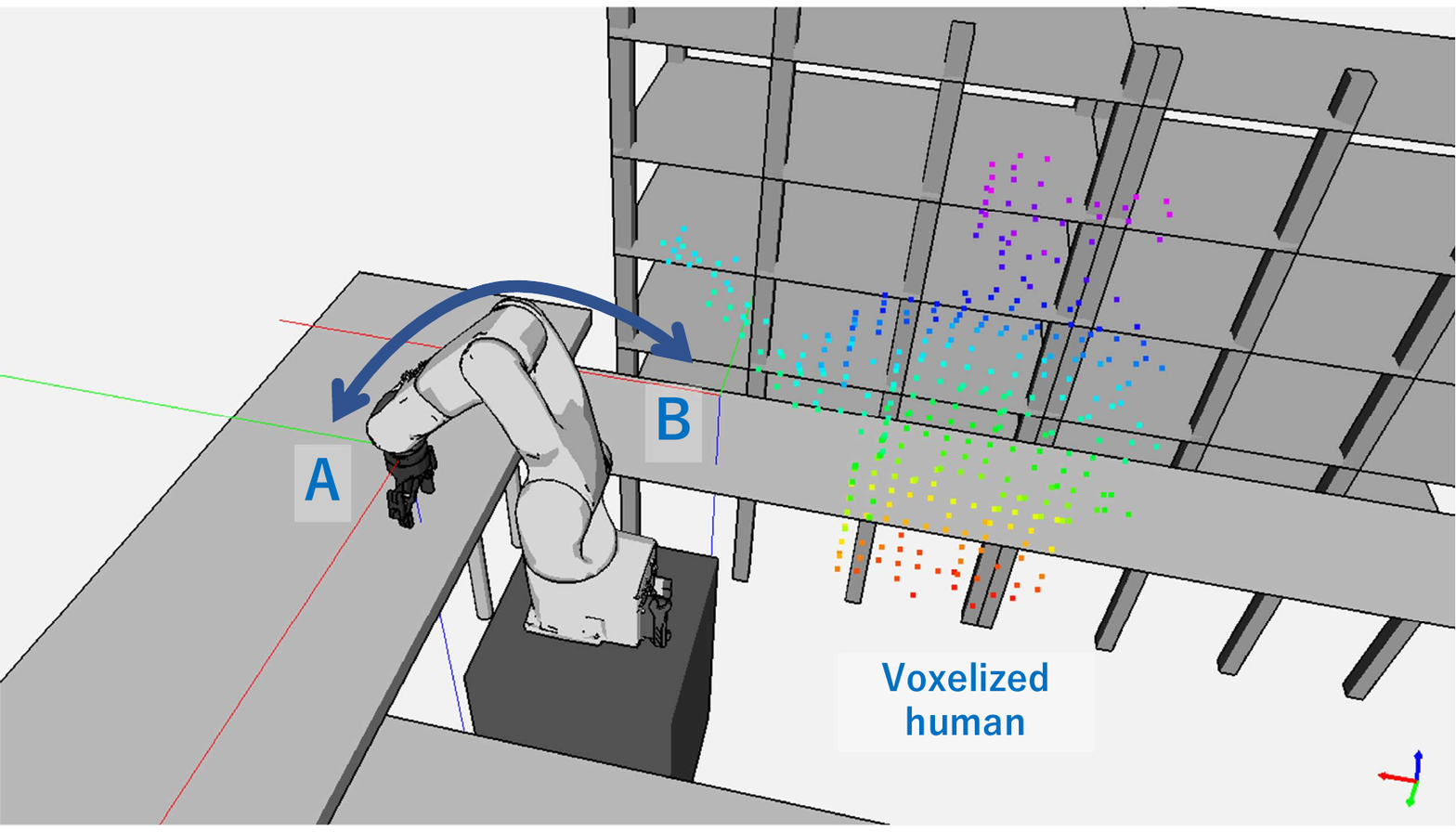}
  \caption{The experimental setup used to compare our method with a simple method which models a robot with spheres.
  The robot loops between point A and point B while the experimenter virtually picks up objects from a shelf aside the robot hindering the robot's motion.
  }
  \label{fig:comparison_experiment_setup}
\end{center}
\end{figure}

Secondly, we compare our method with sphere-based distance checker in simulation (\cref{fig:comparison_experiment_setup}).
The robot loops between point A and point B while the experimenter is in close proximity to the robot and randomly moves his arms aside the robot impeding the robot's motion.
The robot's motion is planned for each trajectory at runtime.
The experimenter's motion is recorded by a Kinect v2, and we replay the obtained sequence of pointclouds in each experiment at the same timing.
The pointcloud is converted into 4 \si{cm} voxels at runtime.
We record a total time trajectory execution time for the robot to move back and forth for 6 laps over 10 trials.  %
The protective distance $d_{prot}$ is set to $3$ \si{cm} and the extent of Link SDFs $\mathbf{e_r}$ is set to $1.2$ m.
The resolution of pre-computed Link SDFs is set to 1 \si{cm}.
The clearance threshold is set to $sqrt((4/2)^2\times3) + sqrt((1/2)^2\times3) \approx 4.3$ \si{cm}.
Our code is based on OpenRAVE~\cite{diankov_thesis}.

At runtime, after planning a trajectory between points using an off-the-shelf RRT-based motion planner in OpenRAVE and before executing the trajectory, intermediate waypoints are sampled using TOPP-RA's automatic gridpoint suggestion feature~\cite{phamNewApproachTimeOptimal2018}.
The number of waypoints (i.e. the batch size) ranges in 300--500 depending on each trajectory.
Robot SDFs are then computed with our proposed method for each waypoint configuration in a parallel, batched manner.
During the trajectory execution, the computed SDFs are used to retrieve the distances between a robot and obstacles for each waypoint and time-optimal safe velocity is computed and applied to a robot at every control cycle based on ~\cite{fujii2023timeoptimal}.

To ensure the safety, $\mathbf{e_r}$ needs to be large enough to capture the obstacle coming closer to a moving robot.
We select the value (1.2 \si{m}) as follows:
Given the joint velocity limit $v_{limit, i}$ and acceleration limit $a_{limit, i}$ for joint i,
the maximum braking time $t_{brake}$ is computed as ${\displaystyle \max_i{\frac{v_{limit, i}}{a_{limit, i}}}}$, which is 0.2 sec for DENSO Robot VS-060.
Therefore, given the maximum velocity of obstacles $v_{obs}$, the system needs to capture obstacles coming closer less than $v_{obs}t_{brake} + d_{prot}$ from a robot trajectory.
In our case, considering the maximum size from the center of robot links is 0.6 \si{m}, $\mathbf{e_r}$ must be larger than $1.6 \cdot 0.2 + 0.03 + 0.6 = 0.95$ \si{m}.

\begin{figure}[ht]
  \vspace{-2mm}
  \begin{center}
  \includegraphics[width=0.70\hsize,trim={0cm 0cm 0cm 0cm},clip]{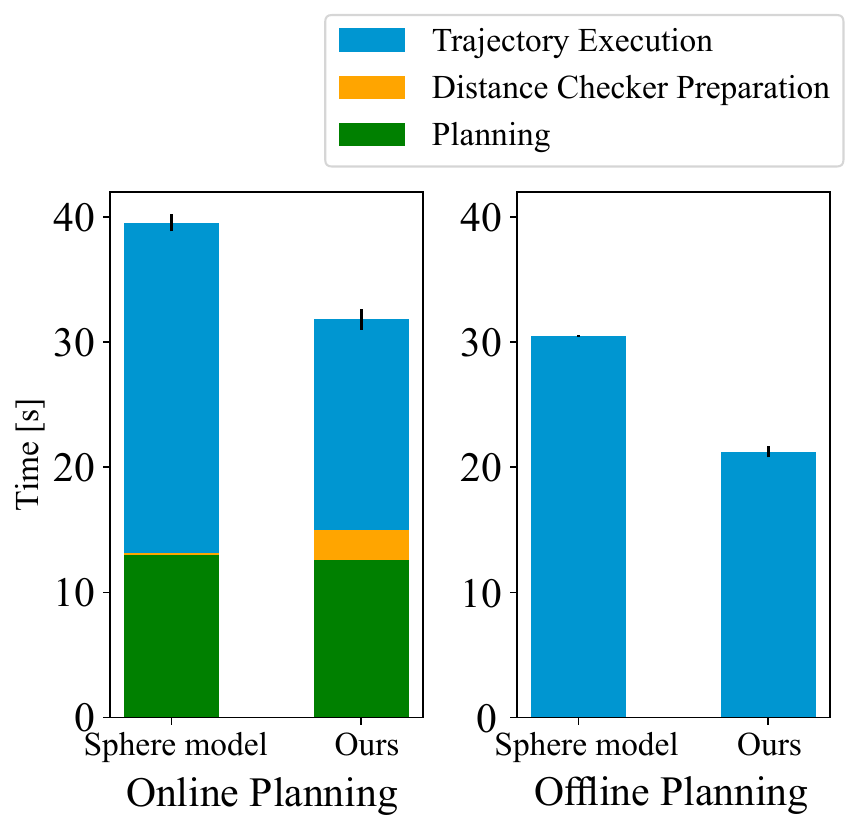}
  \caption{
    Comparison of execution times in a dynamic environment.
    Despite the additional time spent on preparing SDFs, our method exhibits a 1.24x faster total execution time compared to the simple sphere model.
    In case of offline planning (i.e. trajectories are fixed and preparation is done offline), we observe a 1.44x speedup.
  }
  \label{fig:comparison_execution_time}
  \vspace{-4mm}
  \end{center}
\end{figure}

\begin{figure*}[htbp]
  \begin{center}
    \includegraphics[width=0.90\linewidth,trim={2cm 0cm 2cm 0.0cm},clip]{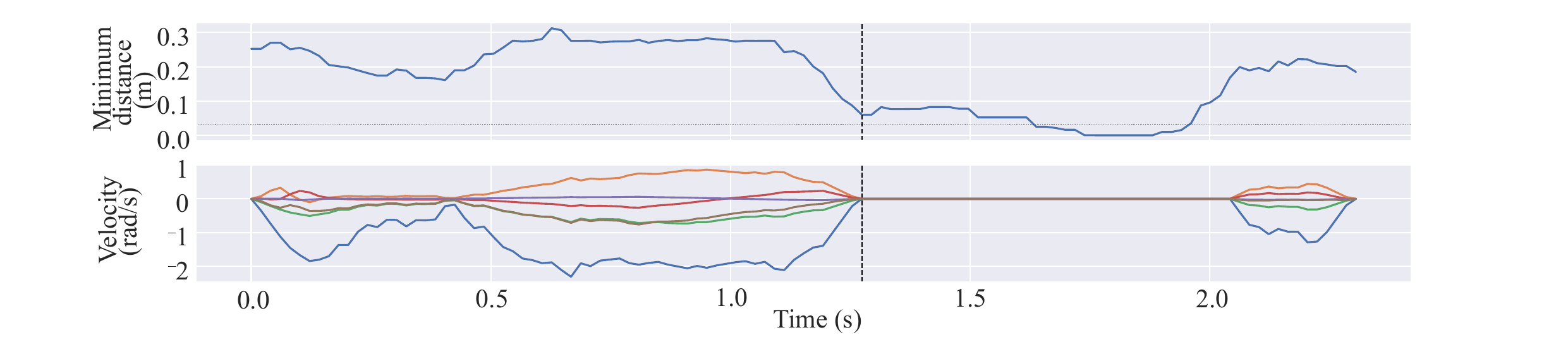} %
    \caption{A result of a trajectory execution, logging Minimum distance from voxelized obstacles and Joint Velocities.
    }
    \label{fig:linksdf_timeseries_plot}
  \end{center}
  \vspace*{-3mm}
\end{figure*}

\begin{figure}[tbhp]
  \centering
  \begin{minipage}[b]{0.99\linewidth}
    \centering %
    \includegraphics[width=0.99\hsize,trim={2cm 4cm 4cm 11cm},clip]{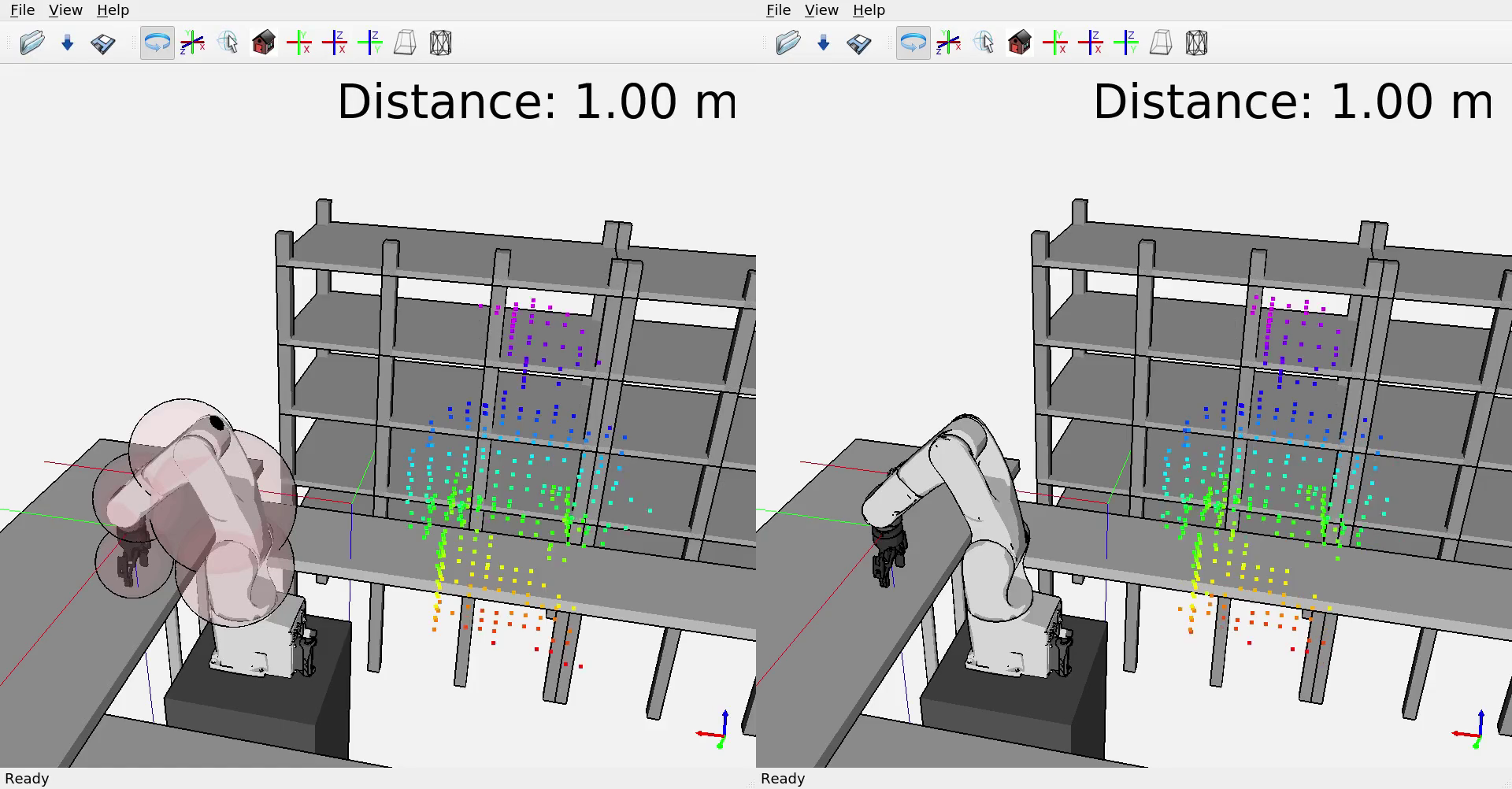}
    \vspace{-6mm}
    \subcaption{}
    \label{fig:robotexperiment_clip3}
  \end{minipage}
  \begin{minipage}[b]{0.99\linewidth}
    \centering %
    \includegraphics[width=0.99\hsize,trim={2cm 4cm 4cm 11cm},clip]{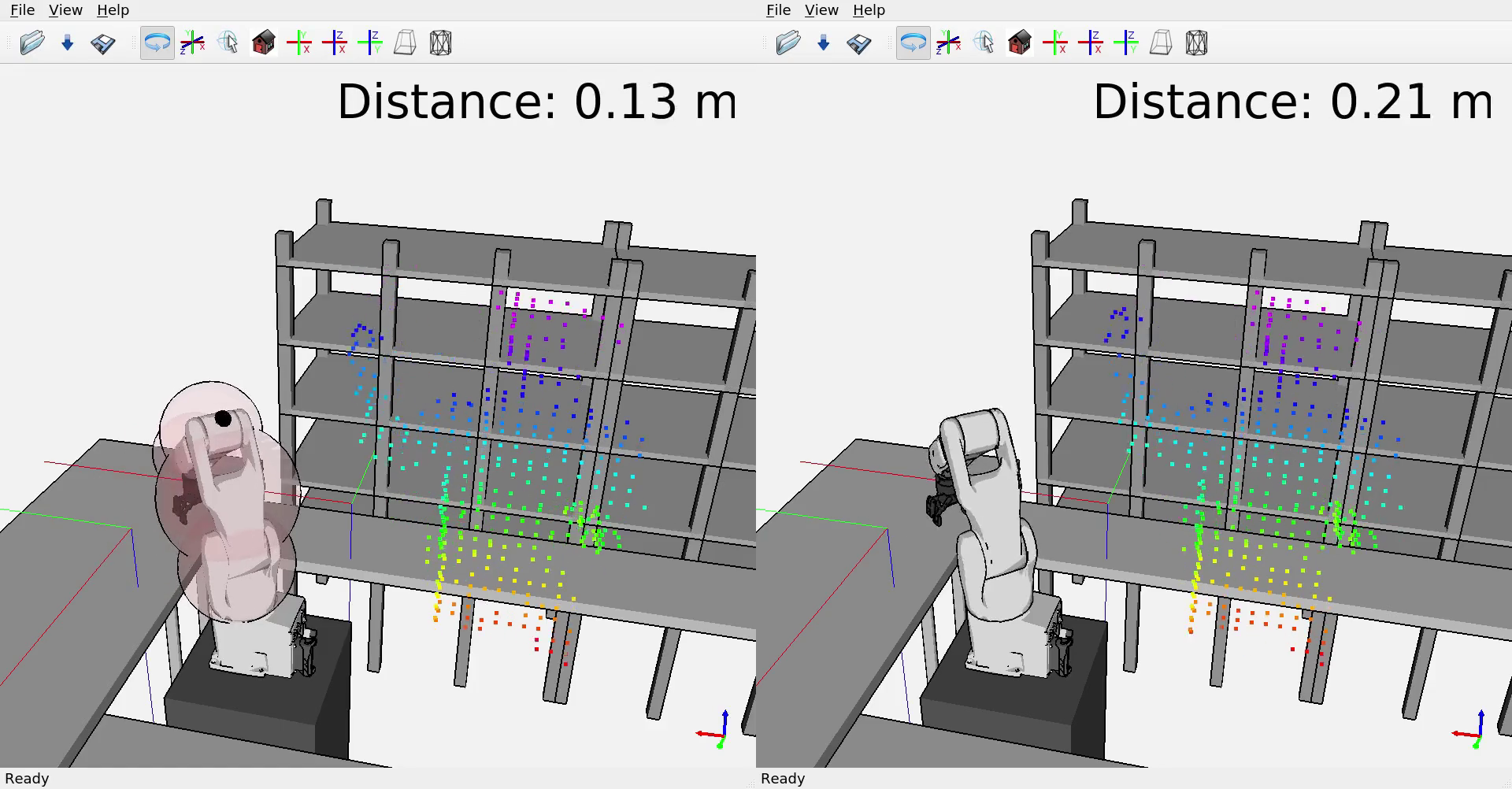}
    \vspace{-6mm}
    \subcaption{}
    \label{fig:robotexperiment_clip4}
  \end{minipage}
  \begin{minipage}[b]{0.99\linewidth}
    \centering %
    \includegraphics[width=0.99\hsize,trim={2cm 4cm 4cm 11cm},clip]{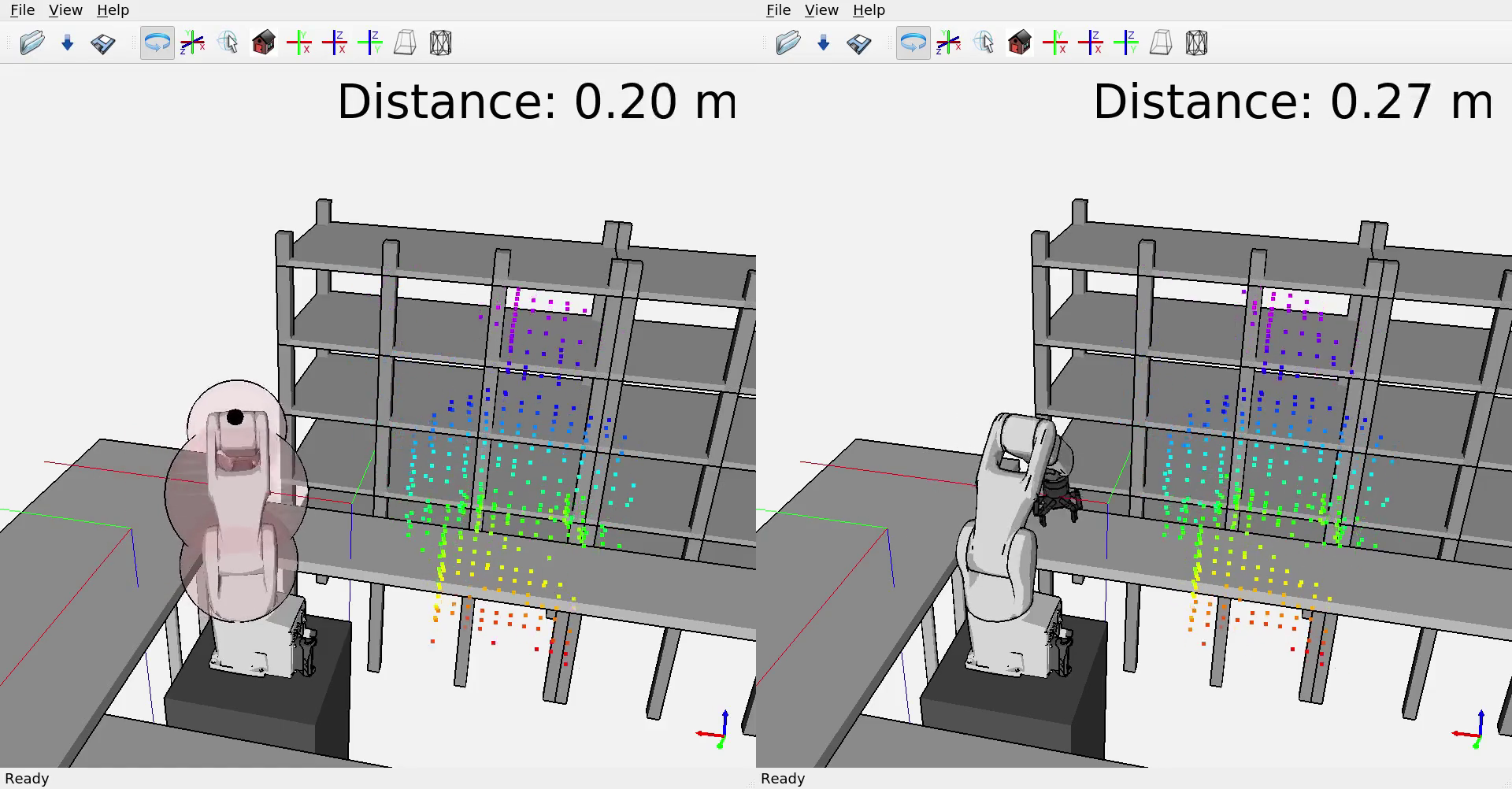}
    \vspace{-6mm}
    \subcaption{}
    \label{fig:robotexperiment_clip5}
  \end{minipage}
  \begin{minipage}[b]{0.99\linewidth}
    \centering %
    \includegraphics[width=0.99\hsize,trim={2cm 4cm 4cm 11cm},clip]{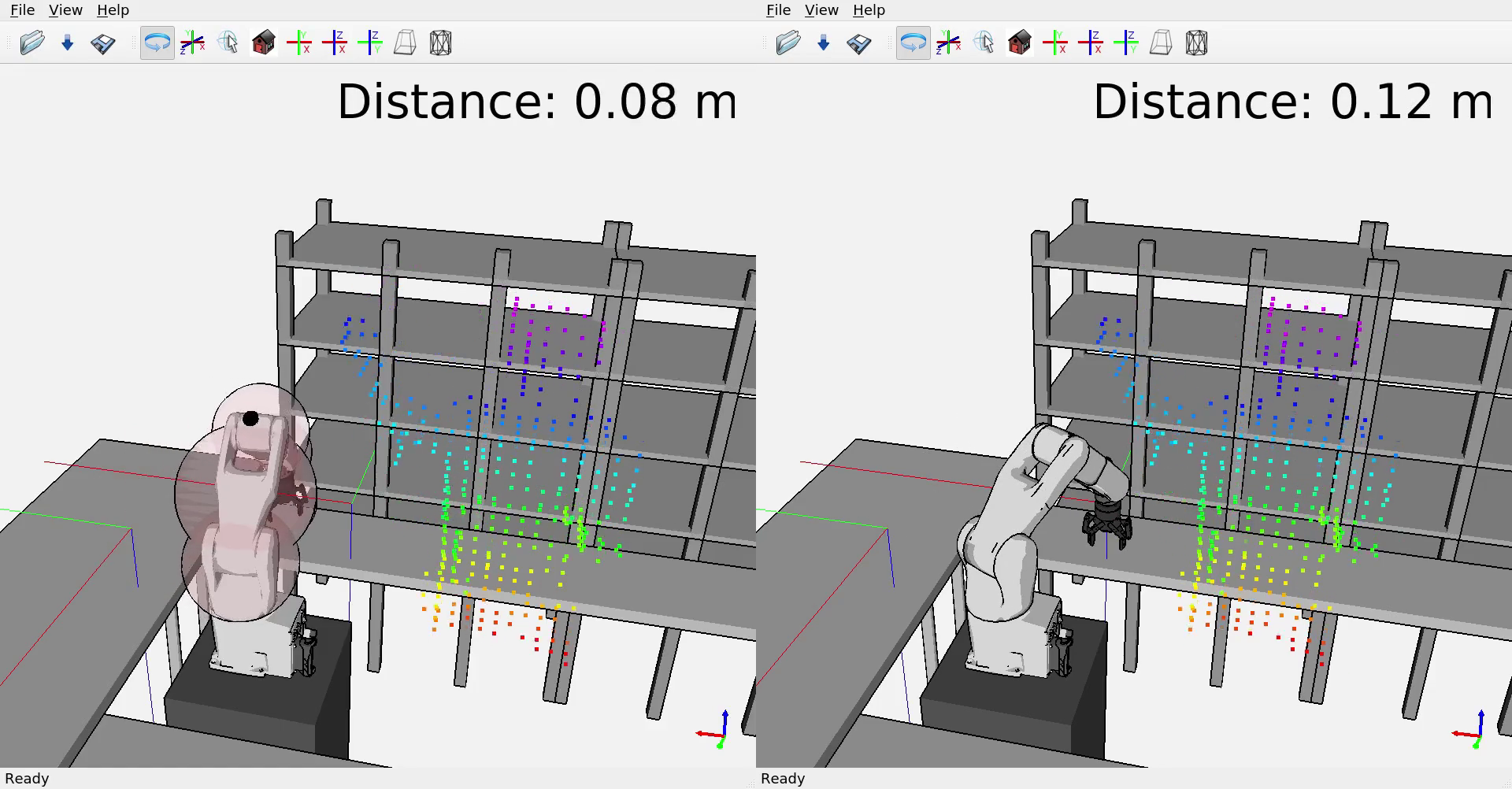}
    \vspace{-6mm}
    \subcaption{}
    \label{fig:robotexperiment_clip9}
  \end{minipage}
  \caption{
    Experimental comparison of our method with a simple sphere robot model.
    Left: Sphere model / Right: Our method.
    The same recording of the experimenter's pointcloud is applied to both methods for comparison.
    The captured video clips show the first trajectory execution.
    (a) The robots start to move almost simultaneously. Ours is a little slower due to SDFs computation.
    (b) Both robots move almost at the same speed initially and starts being hindered by the experimenter.
    (c) Ours accelerates earlier,
    (d) arrives at the destination earlier. %
    The full experiment can be viewed at \url{https://youtu.be/wO6PiOlsu-w}.
  }
  \label{fig:robotexperimentclips}
\end{figure}

As a result, the robot with sphere model takes 39.53 \si{sec} to execute its entire task while ours takes 31.81 \si{sec}, which is 1.24 times faster (\cref{table:execution_time_comparison}, \cref{fig:robotexperimentclips}).
The SDFs computation takes approximately 0.2--0.3 \si{sec} per one trajectory, which is reasonable to compute at runtime.
During trajectory execution, the sphere-based checker takes 5.47 \si{\milli \second} to batch-process distances even on GPU, while our batched distance checker requires only 0.4 \si{\milli \second}.
This is beneficial for achieving high-throughput in a real-time control system (\cref{fig:comparison_execution_time}, \cref{table:execution_time_comparison}).
It is worth noting that our method invests 2 seconds in total at runtime in SDFs preparation for each trajectory execution and is still faster.
If trajectory replanning is unnecessary and the robot follows fixed trajectories and SDFs computation can be done in advance, the total improvement is 1.44 times (from 30.51 \si{\second} to 21.25 \si{\second}).
The comparison videos can be checked from \url{https://youtu.be/wO6PiOlsu-w} and \url{https://youtu.be/YBPpki4fGF8}.
\cref{fig:linksdf_timeseries_plot} illustrates a typical trajectory execution, logging the minimum distance to obstacles and joint velocities.
A dotted horizontal line in the plot represents the protective distance.
This plot shows that the safety is secured by stopping the robot before collision happens.

\begin{table}[ht]
\begin{center}
\vspace{-2mm}
\caption{Comparison of execution times per one trajectory}
\begin{tabular}{lll}
                                                                                                     & \begin{tabular}[c]{@{}l@{}}Sphere\\ model\end{tabular} & \begin{tabular}[c]{@{}l@{}}Ours\\ (neural approx.)\end{tabular} \\ \hline
\begin{tabular}[c]{@{}l@{}}Distance Checker Preparation \\ per 1 trajectory [\si{\second}]\end{tabular}                    & $0.15 \pm 0.012$                                       & $2.34 \pm 0.057$                                                \\
\begin{tabular}[c]{@{}l@{}}Runtime Evaluation \\ per 1 trajectory [\si{\milli \second}]\end{tabular} & $5.47  \pm 0.096$                                      & $0.391 \pm 0.0014$                                              \\
\hline
\end{tabular}
\label{table:execution_time_comparison}
\vspace{-2mm}
\end{center}
\end{table}

\subsection{Real Robot Experiment} \label{subsec:linksdf_realrobot_experiment}
Finally, we experimentally test our algorithm with a real 6 DoF robot.
During the experiment, the robot moves between several configurations while an experimenter inhibits its motion.
The environment is captured using Intel RealSense\texttrademark D455 and voxelized with a resolution of 4 \si{cm}.
The maximum speed of obstacles is set to 2.0 \si{\meter / \second}, and the robot speed is limited to 80\% of its capacity for the safety of the experimenter.%
The experiment can be viewed at \url{https://youtu.be/iZP_6rD34lA}.

\section{Conclusion} \label{sec:conclusion}
A batched, fast and precise distance checker has been required for truly time-optimal safe path tracking in human-robot collaborative environments.
In this paper, we propose a real-time batched distance checking method based on precomputed link-local SDFs.
Our method can check distances between a robot and obstacles along a trajectory within less than 1 millisecond on GPU, which is suitable for time-critical safety control under a collaborative situation.
Additionally, to accelerate a preprocessing process, a neural approximation has been proposed, which makes the preprocessing 2x faster.
Finally, we have experimentally demonstrated that our method can navigate a robot earlier than a fast but conservative geometric-primitives based distance checker in a dynamic environment.

It should be mentioned that, by introducing a greater number of smaller spheres in the simple sphere model to enhance its precision, the performance improvement of our method will diminish and the time invested in SDFs preparation may not be justified.
However, as the number of spheres and occupied voxels increases, the runtime speed of the sphere-based checker grows linearly, which is critical for ensuring safe control.
Indeed, during our experiments, we encountered situations where the runtime speed was insufficient for the control cycle, particularly in larger and congested environments with numerous occupied voxels.
On the other hand, our method maintains an advantage in terms of runtime speed.
Although it scales linearly to the number of occupied voxels, it remains significantly faster since it only requires GPU memory access at runtime.

One of the limitations of our approach is its scalability in terms of space complexity.
It is not well-suited for a large, finely-voxelized environment due to the GPU memory consumption associated with cached SDFs.
Although we do not observe a significant memory overhead inherently as the environment size increases, the GPU memory required to store SDFs increases linearly with the number of waypoints and the number of environment voxels.
For instance, during the experiment described in \cref{subsec:sim_robot_comparison}, about 3GB GPU memory was consumed at runtime.
While this issue can be mitigated by employing multiple GPUs, but considering the hardware cost, further work is needed to reduce memory consumption.

\bibliographystyle{unsrt}
\bibliography{root}

\end{document}